# HANDLOOM DESIGN GENERATION USING GENERATIVE NETWORKS

*Rajat Kanti Bhattacharjee*⋆   *Meghali Nandi*⋆   *Amrit Jha*⋆

Gunajit Kalita⋆   Ferdous Ahmed Barbhuiya   †

⋆ Assam Engineering College, Guwahati, India
† Indian Institute of Information Technology Guwahati, India

## ABSTRACT

This paper proposes deep learning techniques of generating designs for clothing, focused on handloom fabric and discusses the associated challenges along with its application. The capability of generative neural network models in understanding artistic designs and synthesizing those is not yet explored well. In this work, multiple methods are employed incorporating the current state of the art generative models and style transfer algorithms to study and observe their performance for the task. The results are then evaluated through user score. This work also provides a new dataset "NeuralLoom" for the task of the design generation.

## 1. INTRODUCTION

The fashion trends have always been user-specific, based on their likes and taste. In such a scenario, traditional Handloom of ethnic culture suffers the most, with the industry being an occasional tourist attraction. Variation in the design can bring a new life to this industry but bringing out such variations is difficult to achieve especially by the people currently in the industry. The human mind is conditioned to bias and coming up with designs that could incorporate the style of non-Handloom designs on the Handloom designs or vice versa is a tough task. Quantitatively defining such merging of design style and the taste is difficult in scientific terms. Hence generating design or texture from existing prior or randomly sampled priors, which are agnostic to both traditional design and also the more generic designs is what this particular work will try to address. This problem can also be framed as "Transfer or Generation of texture and geometric artefact in an image", where the transfer will mean establishing the design of a Handloom on existing generic designs. Generation means the generation of something completely original given suggestion via some random noise. For this, the idea of design and texture need to be established which are discussed in the subsequent sections.

### 1.1. Image texture, content and style

Texture[1] can be considered as a repeated random or deterministic sampling in image spatial domain of a particular organization or cluster of color pattern and gradients. If it is assumed that in any image the geometric structure in it provides the semantics of the image and the color space provides the texture, it wouldn't be hard for humans to render the actual image in their mind, whereas for to program a machine to generate such a coherent RGB combination will be a difficult task. In the literature also it still remains dependent on fixed feature extraction [2] or by using pre-trained neural network for texture transfer i.e Neural Style Transfer[3]. This work was suggestive of preserving content and transferring texture which the author collectively calls as the style of an image. But design remains a much more abstract concept than that.

### 1.2. Design in an Image and its Generation

Design is a much more abstract concept than style. It can be understood by looking at Fig 1. The fig.1 gives a brief idea of how design itself is a complex idea. Though the style can describe the design of content in an image but taking this uniform assumption will result in lacking the aesthetic taste in the end results and will be incoherent for the human requirements. A generative model that works to sample data from a modeled distribution can solve this since the generative model doesn't have to satisfy certain generation constraint dictated by strict image features, but rather to the higher-level conditions of task subjection (Good design vs Bad design). This work contributes in design generation task through a study on effective techniques ,a dataset to achieve the task and results of a user acceptance test for computer generated designs. This work generates new sample of "Handloom" design using both the regression-based content and texture transfer and design distribution modeling for cross-domain mapping. In the subsequent sections, section 2 highlights the methods used throughout this work, section 3 explains the "Neural-Loom"[5] dataset and the other datasets used throughout the experiments, section 4 and 5 discusses the methods and significance of the results along with providing insights into the user acceptance scores.

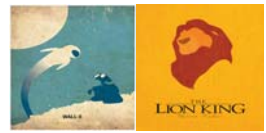

**Fig. 1**: Wall E  Lion King posters [4] Minimalist design with different texture.

## 2. RELATED WORKS

Convolutional Neural Network[6] are proven to be well capable of classification task and understanding natural images. But while capturing the image features and transforming them into deep higher-level features inside the deep convolutional networks, these networks also capture the internal image statistics that dictate how the texture of an image works. This idea was used for style transfer in Neural Style Transfer[7]. It exploited Deep Convolution Neural network's ability to create a hierarchy of information in its layer where the early layers can be used for content transfer and higher layers can be used for texture transfer. This work was then improved upon by turning the optimization-based approach to training a neural network approach in Perceptual Losses for Real-Time Style Transfer and Super-Resolution [8]. Here further improvement was achieved by controlling the factors affecting the image generation by [9]

For the task of design generation the family of Generative Adversarial Network [10] that works to transfer between domains of data was chosen. This method allows the generative network's internal feature representation to explore more of the solution space and hence arrive at better results that are close to the solution distribution samples or is part of the solution space distribution. Deep Convolutional GANs[11] though not working as domain transfer but as a sampling measure from some prior distribution were considered for generating designs. A major problem with the above-mentioned methods was the performance of efficient inference in directed probabilistic models, in the presence of continuous latent variables with intractable posterior distributions, and large datasets. Variational AutoEncoder[12] solved the problem by modeling the autoencoder[13] architecture as probability density function that outputs based on prior distributions but lack of massive dataset and variation in it restricts its usage in this use case. So instead of working with priors from some chosen distribution, the prior can be set to be from the same domain as the target and hence the problem can be framed as an image to image translation problem addressed by [14] which frames the problem as Conditional GANs [15]. However, this translation does not guarantee a one-to-one mapping between the domains as there may be infinitely many mappings that may induce the same distribution. For addressing this Cycle GAN [16] adds more structure to the objective by trying to eliminate the effect of mode collapse by implementing a cycle consistency [17] which should be maintained in the translation of images across domains. This is further improved on by introducing a reconstruction loss at the cycle consistency end in DiscoGAN [18]. One of the recent work Design Inspired Generative Networks [19] takes a similar approach to create design distribution model and put designs on common clothing. Though it deals with a similar subject as this work, Design GAN[19] targets different clothing and apparel which has lesser complex embroidery work than traditional Handloom. This work tries to address the same problem when taken in the context of complex designs.

## 3. DATASET

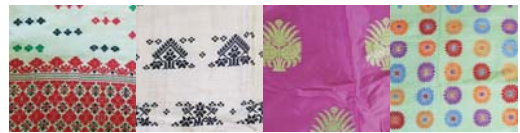

**Fig. 2**: Sample Images of "Regional Handloom" dataset, more at [5]

The data, for the image to image translation and neural style transfer target, was collected by taking photographs of Indian Saree's. A Saree is a common Indian women apparel often possessing complex artistic design on it. The dataset consist of both printed and local Handloom fabric variant having sufficient design complexity from various individual participants. The collection comprised of two classes, namely, the "Regional Handloom" and the "General Handloom" and is termed as "Neural-Loom"[5]. "Regional Handloom" comprises design specific to a geographical location and "General Handloom" comprises of collection of generic designs not specific to any geographic region.

The need for the handcrafted dataset arose because there is no dataset of such kind. Scrapping the internet and e-commerce website for such images would only lead to poor quality and low-resolution images, hence the decision to work towards collecting a custom dataset. The dataset comprises of 33 "Regional Handloom" and 500+ "General Handloom" high-resolution images. The images were reshaped, re-sized and randomly cropped for taking out the artistic design segments out for training the networks. Nine types of augmentation methods such as rotating on different angles and flipping color axis were employed to artificially increase the dataset count when needed. The images if used for edge-to-image translation does not yield good edge results hence it is recommended to use Otsu threshold[20] to create a binary mask of design structure and its background and then use it for the purpose. The dataset has been open-sourced on GitHub for the benefit of the researchers. The dataset used for neural style transfer network was the Microsoft COCO[21] as suggested by the authors [8]. The validation set of COCO was used for the purpose of this work.

## 4. METHOD

For addressing the problem of design generation, it is framed as a problem of texture transfer and texture plus content generation. For creating design data, patches were extracted from the images of "Neural-Loom" dataset where each patch represented some design patterns. These random crops were generally around the resolution of 256x256 or 512x512. This was done with the intent of keeping the sizes uniform and have locally relevant information on designs on a patch. The dataset was also run through some common augmentation methods to artificially increase the size of the dataset for the learning task by the model. Similarly, the COCO dataset was also modified to have similar resolution irrespective of the scaling and ratio since the focus was to generate designs not relevant to the structure of real-world object and artefacts. For the problem set, the following are the various pipelines that were devised.

- **Generate Design from prior standard distribution:** GANs were the choice of network for this purpose. The design images are used to train a DCGAN and Variational Autoencoder. It turned out that basic DCGAN or VAE were not able to give any noticeable usable results. It's unknown as to why during training the DCGAN kept collapsing and giving noise as output, was it because of the dataset size or the model architecture was not appropriate for the task is something that would need further investigation.

- **Texture transfer from some trained prior on target:** This approach took an image of a "General Handloom" design from "Neural-Loom" (target content) and fused it with the texture or style of some other "Regional Handloom" over which the generator network is trained on. The target design here is first converted to grayscale and then converted to a binary mask using the Otsu threshold to separate out the background and the foreground of the target. The background and foreground masks can now be used to mask and paint the background and foreground of the target image separately with separate styles from two prior separate Handloom style each generated from the neural style transfer network. The neural style transfer network for this experiment was trained using COCO dataset as the input and the various design from the "Neural-Loom" dataset as the target style. This allowed transferring the style of the Handloom on the target input during inference.

- **Design fusion by domain transfer :** Though the vanilla GANs failed CycleGAN and DiscoGAN showed better results. With CycleGAN two experiments were devised "NormalSaree2Handloom and "CoCo2Handloom".

    **CoCo2Handloom:** The experiment was inspired by the result of CycleGAN's results of "Ramen2People"[22]. The objective was for the network to learn a mapping between the domain of natural images and design images. The experiment was repeated for both "Generic Handloom" and "Regional Handloom" from the dataset.

    **Saree2Handloom:** This experiment is similar to the one before and targets generating "Regional Handloom" from "Generic Handloom" samples.

- **Hand drawn mask to design** This experiment uses DiscoGAN[18] to generate design from the given hand-drawn artefact which is in the form of binary masks. This application specifically focused on having designers full control over the kind of artefacts and shapes present in the outcome but have in-painting that is specific to any design style the network is trained for. The training was done using the collection of all "Neural Loom" dataset.

All of the various methods were bundled in a usable web application[23] that could demonstrate the above-mentioned approach's capabilities. The application is currently hosted over Github as open access. The results from the above methods are presented in section 5

## 5. RESULTS

As discussed in the above section 4, several approaches were chosen to be tried and tested for the task in hand. The results of the different methods taken up are discussed below.

- **Texture transfer from some trained prior on target:** The results obtained from training Fast Neural Style Transfer on prior selected Handloom style can be seen in Fig. 3.

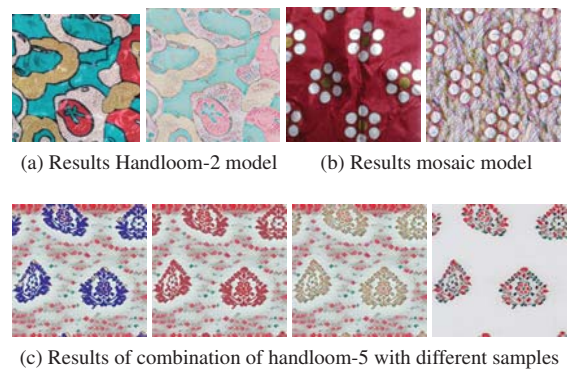

(a) Results Handloom-2 model    (b) Results mosaic model

(c) Results of combination of handloom-5 with different samples

**Fig. 3**: The results are based on models pre-trained on various abstract design and patterns from "Neural-Loom". The results are generated from the final prototype application[23]

- **Generic Handloom to Regional Handloom (Cycle-GAN):** The results obtained from the Cycle GAN model for "Generic Handloom" to "Regional Handloom" translation can be seen in Fig. 4. The results are not promising given only change in color is observed which heavily reflects the bias of the "Regional Handloom" in the dataset samples due to its low count and samples with lighter color tone.

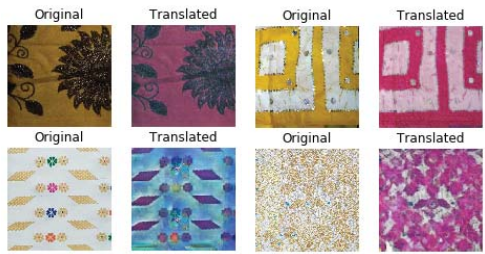

**Fig. 4**: CycleGAN results for Generic Handloom to Regional Handloom model

- **Natural Images to Handloom (CycleGAN):** The results obtained from the Cycle GAN model for natural images to Handloom translation can be seen in Fig. 5.

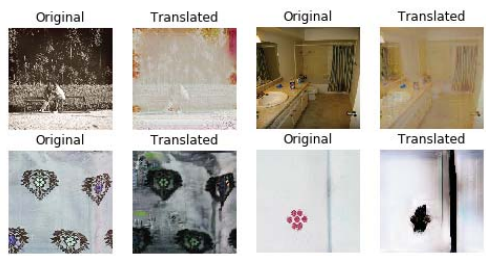

**Fig. 5**: CycleGAN results for NaturalImages-to-Handloom model

- **Hand-drawn mask to design** The results obtained from the DiscoGAN model for coloring hand-drawn images can be seen in Fig. 6. The results shown here gives promising direction towards having users or designers have fine-grain control in what kind of design one wants with the freedom of combining with any sort of aesthetic style they choose to in-paint their drawings

The results from DiscoGAN and Neural Style Transfer were the most promising ones, with both being used for the final prototype application that enables one to experiment with these methods. The result gives us the insight that such methods can bring about effective ways of creating fusion designs, but a more proper definition of design must be made which need to be incorporated in the solution. The dataset also needs more samples in it to remove any intrinsic bias and to have a more vivid and varied dataset to work with. Using the results generated from the application a user review was conducted among 53 participants where there were 35 male and 18 female. Each participant was given a series of samples patch images, which consisted of both real and generated designs. The participants were asked to label the images as "Real", "Generated" or "Not Sure" and rating the design with "Good", "Bad" and "Maybe". It was observed that **45.8%** of users rated the "Generated" design as **"Good"**, 29.7% rated "Bad" and 24.5% were not sure. From the participant's labels7 we can also conclude about users confusion and lesser certainty with generated designs. Which means that such generated design can be really useful direct production or even allow for quicker design iterations for designers.

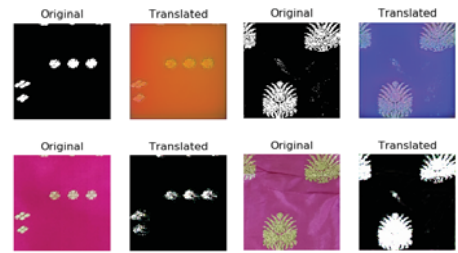

**Fig. 6**: DiscoGAN results

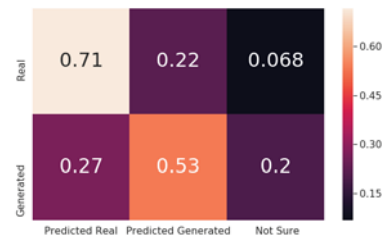

**Fig. 7**: Participants labeling percentages. vertical axis is type of image. horizontal is prediction for same.

## 6. CONCLUSION

Handloom was of special interest due to its intrinsic complexity in design and the socioeconomic impact of it. This paper proposes a study of techniques for the generation of complex handloom design. This paper also contributes through the creation of a new dataset "Neural-Loom" for design generation task and insight into user reception of such computer-generated designs which was mostly positive. The final result was deployed in an application which can act as a creative tool aiding designers. Although the problem of design generation still remains at large an open-ended research area which should be explored further and can lead to a better understanding of the generation system as evident from the study presented in this work.